\documentclass[conference,10pt]{IEEEtran}

 \usepackage[dvips]{graphics,graphicx,epsfig}
 \usepackage[english]{babel}
\usepackage[latin1]{inputenc}

\usepackage{amssymb,amsmath}
\usepackage{url}

\usepackage{cite} 
\usepackage{stfloats}  

\begin{document}

\title{A two-step fusion process for multi-criteria decision applied to natural hazards in mountains}
\author{\IEEEauthorblockN{Jean-Marc Tacnet}
\IEEEauthorblockA{\emph{Cemagref}-ETNA\\
2, rue de la papèterie B.P. 76\\
38402 Saint Martin d'Heres Cedex, France\\
Email: jean-marc.tacnet@cemagref.fr}
\and
\IEEEauthorblockN{Mireille Batton-Hubert}
\IEEEauthorblockA{EMSE - SITE\\
29, rue Ponchardier\\
42100 Saint-Etienne, France\\
Email: batton@emse.fr}
\and
\IEEEauthorblockN{Jean Dezert}
\IEEEauthorblockA{ONERA\\
29, avenue de la division Leclerc B.P. 72\\
92322 Châtillon Cedex, France\\
Email: jean.dezert@onera.fr}
}

\maketitle

\selectlanguage{english}

\begin{abstract}

Mountain river torrents and snow avalanches generate human and material damages with dramatic consequences. Knowledge about natural phenomenona is often lacking and expertise is required for decision and risk management purposes using multi-disciplinary quantitative or qualitative approaches. Expertise is considered as a decision process based on imperfect information coming from more or less reliable and conflicting sources. A methodology mixing the Analytic Hierarchy Process ($AHP$), a multi-criteria aid-decision method, and information fusion using Belief Function Theory is described. Fuzzy Sets and Possibilities theories allow to transform quantitative and qualitative criteria into a common frame of discernment for decision in \emph{Dempster-Shafer }Theory ($DST$) and \emph{Dezert-Smarandache} Theory ($DSmT$) contexts. Main issues consist in  basic belief assignments elicitation, conflict identification and management, fusion rule choices, results validation but also in specific needs to make a difference between importance and reliability and uncertainty in the fusion process.

\end{abstract}


\IEEEpeerreviewmaketitle

    \section{Introduction}

\begin{figure}[htb]
         \begin{center}
        \includegraphics[width=88mm]{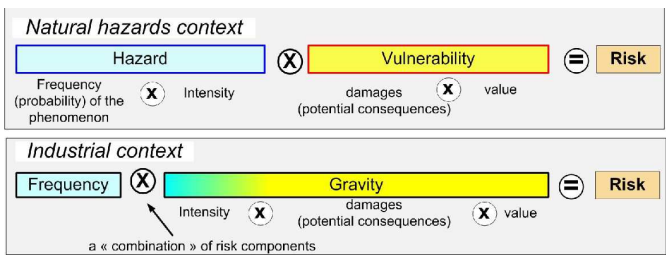}
        \end{center}
        \caption{Equations of risk in natural hazards and industrial contexts.}
        \label{fig:FormulationRisqueGB}
    \end{figure}

    Mountain river torrents and snow avalanches generate human and material damages with dramatic consequences. In the natural hazards context, risk is assessed as a combination of  hazard and  vulnerability levels. This formulation can be considered as equivalent to a combination of frequency and gravity which is more currently used in  industrial context (figure~\ref{fig:FormulationRisqueGB}).

\begin{figure}[htb]
        \begin{center}
        \includegraphics[width=88mm]{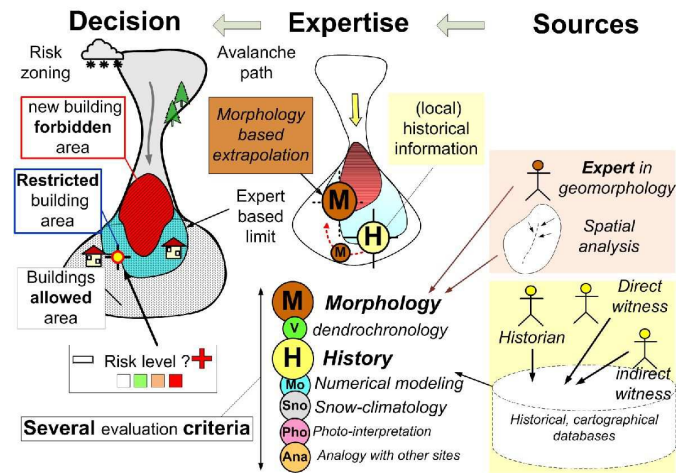}
        \end{center}
        \caption{Information, expertise and decision in risk zoning applications.}
        \label{fig:PPRFusionGB}
     \end{figure}

     Expertise is always required to define the types of possible phenomena, to assess the hazard and risk levels and to propose prevention measures. Expert judgements depend on quality and uncertainty of the available information that may result from measures, historical analysis, testimonies but also subjective, possibly conflicting, assessments done by the experts themselves. As an example, the definition of risks zones is often based on the extrapolation of historical information known on particular points using morphology based analysis (figure~\ref{fig:PPRFusionGB}).

    At the end, phenomenon scenarios and decisions may very well rely on very uncertain information without being able to really know what was completely true, imprecise, conflicting or simply unknown in the hypotheses leading to these results. In that context, our essential hypothesis consists in considering expertise as a decision process based on imperfect information related to multiple criteria and coming from more or less reliable and conflicting sources.

    This paper proposes a methodology able to help decision based on imperfect information. In section ~\ref{sec:Methods}, we briefly introduce the principles of multi-criteria decision analysis ($MCDA$) focusing on the $AHP$ developed by T. Saaty (section  ~\ref{sec:AHP}). The section ~\ref{sec:DecisionImperfection} analyzes the existing methods methods using both $MCDA$ methods and theories for uncertainty management. In section ~\ref{sec:ErMcdaMethodology}, we present the different steps of a new methodology applying to a multi-criteria decision problem based on imperfect information resulting from more or less reliable sources. Conclusions and perspectives are given in section ~\ref{sec:Conclusion}.

    \section{Methods for Multi-criteria Decision analysis and imperfect information}

     \label{sec:Methods}

    Information and decision are closely linked and different methods exist to take a decision on the basis of imperfect information. From one hand, main principles of multi-criteria decision analysis and existing theories to manage imperfect information are over-viewed. From the other hand,  we then briefly analyze the characteristics and lacks of existing methods methods using both $MCDA$ methods and theories for uncertainty management.

        \subsection{The Analytic Hierarchy Process }

    \label{sec:AHP}

\begin{figure}[htb]
        \begin{center}
        \includegraphics[width=88mm]{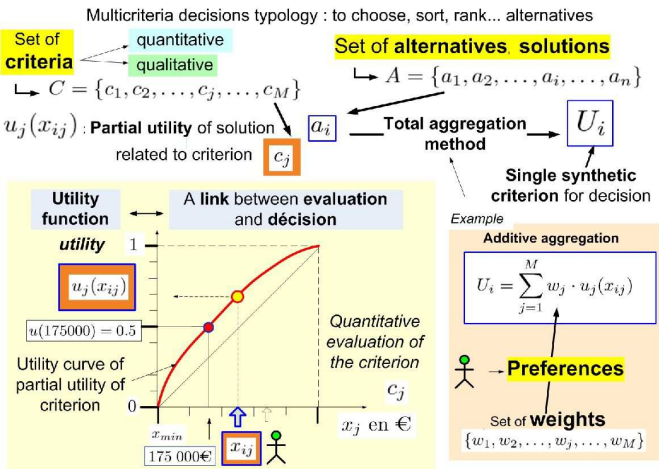}
        \end{center}
        \caption{Multi-criteria decision method based on a total aggregation principle.}
        \label{fig:PbmatiqueMAUTGB}
   \end{figure}

    Multi-criteria decision analysis aims to choose, sort or rank alternatives or solutions according to criteria involved in the decision-making process. Main steps of a multi-criteria analysis consist in identifying decision purposes, defining criteria, eliciting preferences between criteria, evaluating alternatives or solutions and analyzing sensitivity with regard to weights, thresholds, \ldots. Total aggregation methods such as the Multi-Attribute Utility Theory (M.A.U.T.)~\cite{Keeney1976, Dyer2005} synthesizes in a unique value the partial utility related to each criterion and chosen by the decision maker. Each partial utility function transforms any quantitative evaluation of criterion into an utility value. The additive method is the simplest method to aggregate those utilities (figure~\ref{fig:PbmatiqueMAUTGB}).

\begin{figure}[htb]
        \begin{center}
        \includegraphics[width=88mm]{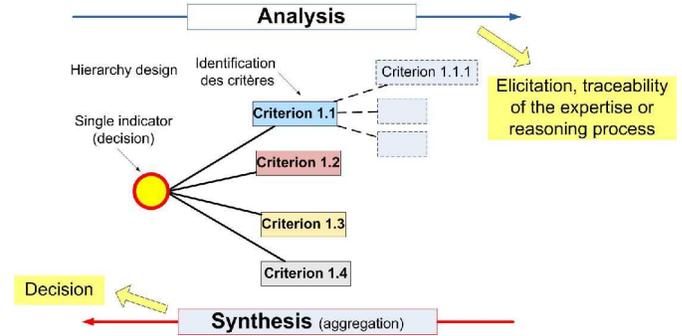}
        \end{center}
        \caption{Principles of the analytic hierarchy process.}
        \label{fig:AnalyseSyntheseGB}
    \end{figure}

    The Analytic Hierarchy Process ($AHP$) ~\cite{Forman2002,Saaty1980,Saaty2005} is a single synthesizing criterion approach. This method is world-wide used in almost all applications related with decision-making \cite{Vaidya2006}. $AHP$ is a special case of complete aggregation method based on an additive preference aggregation and can be considered as an approximation of multi-attribute preference models ~\cite{Dyer2005}. Its principle is to arrange the factors considered as important for a decision in a hierarchic structure descending from an overall goal to criteria, sub-criteria and finally alternatives in successive levels. It is based on three basic steps: decomposition of the problem, comparative judgments and hierarchic composition or synthesis of priorities (figure~\ref{fig:AnalyseSyntheseGB}).

\begin{figure}[htb]
        \begin{center}
        \includegraphics[width=88mm]{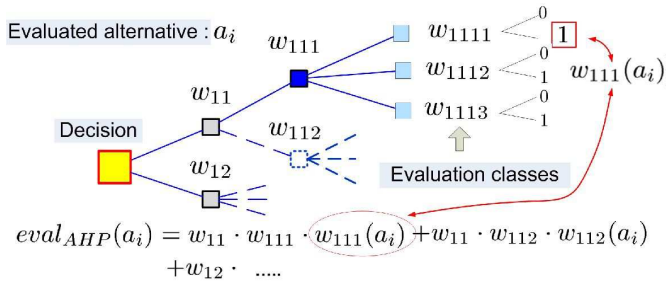}
        \end{center}
        \caption{Criteria-Estimator-Solution model.}
        \label{fig:AhpPrinciplesGB}
    \end{figure}

    At each level of the hierarchy, a preference matrix is built up through pairwise comparisons using a semantic and ratio scale to assess the decision maker preferences between the criteria of the considered level. Through the $AHP$ pairwise comparison process, weights and priorities are derived from a set of judgments that can be expressed either verbally, numerically or graphically. The original $AHP$ process uses an additive preference aggregation and compares the solutions from one to each other in a so-called "Criterion-alternative approach". This implies to make pairwise comparisons between all the solutions or alternatives in order to obtain preferences levels between these alternatives. When dealing with great amount of data, this becomes quickly quite difficult. An other approach so-called "Criterion-index (or estimator)-alternative" is used in our developments (figure~\ref{fig:AhpPrinciplesGB}). Instead of comparing all the alternatives, the decision analyst evaluates, for each alternative, criterion according pre-existing classes. Each evaluation class corresponds to an increasing or decreasing level of satisfaction of a given criterion involved in the decision making.  For example, the criterion \emph{human vulnerability} exposed to natural hazards can be assessed according to three classes based on a number of existing and exposed buildings (figure~\ref{fig:ArbreSimpleGB}). These classes code some kind of ordinal levels corresponding to a low, medium or strong contribution (or satisfaction) to (or of) the criterion. In that way, the $AHP$ method, despite the known issues of complete aggregation methods, fits quite well to decision ranking problems where the alternatives are not all known.

        \subsection{Making a decision on the basis of imperfect information}
        \label{sec:DecisionImperfection}

    Several theories have been proposed to handle different kinds of imperfect information: \emph{Fuzzy Sets Theory} for vague information ~\cite{Zadeh1965}, \emph{Possibility Theory} for uncertain and imprecise information ~\cite{Zadeh1978,Dubois1988} and \emph{Belief Function Theory} that allows to consider uncertain, imprecise and conflicting information. In addition to original \emph{Dempster-Shafer} theory ($DST$) ~\cite{Shafer1976}, \emph{Dezert-Smarandache} ($DSmT$) theory has proposed new principles and advanced fusion rules to manage conflict between sources ~\cite{DSmTBook1-3,Dezert2009}.

      Uncertainty and imprecision in multi-criteria decision models has been early considered ~\cite{Roy1989}. $MAUT$ in general ~\cite{Wang2006b} and $AHP$ in particular have already been associated to the Evidence Theory ~\cite{Beynon2000, Beynon2002} including the cases of several sources ~\cite{Beynon2005a}. A methodology called $ER-MCDA$ (\textbf{E}vidential \textbf{R}easoning - \textbf{M}ulti-Criteria \textbf{D}ecision \textbf{A}nalysis) mixing multi-criteria decision analysis and information fusion ~\cite{Tacnet2009a,Tacnet2009b} is proposed to help the experts in a context of imperfect information. Its main principles are described in section ~\ref{sec:ErMcdaMethodology}.

    \section{The $ER-MCDA$ methodology}

     \label{sec:ErMcdaMethodology}

\begin{figure}[h]
        \begin{center}
        \includegraphics[width=88mm]{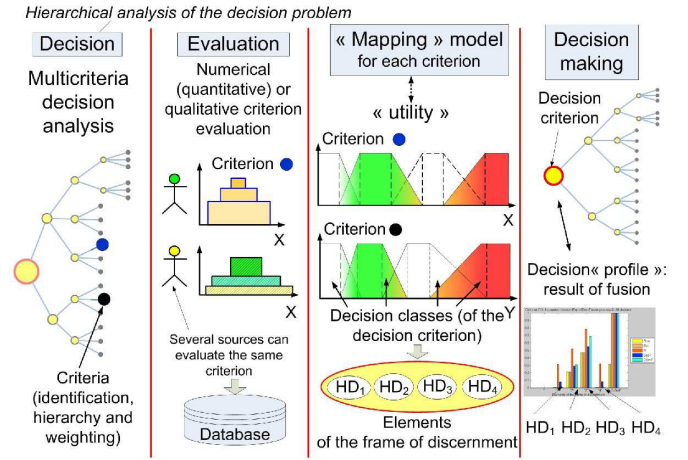}
        \end{center}
        \caption{Four dissociated steps of the $ER-MCDA$ methodology.}
        \label{fig:DissociationGB}
    \end{figure}

    The principle of the $ER-MCDA$ methodology is to use the multi-criteria decision analysis framework to analyze the decision problem and to identify the criteria involved in decision.  Utility functions and aggregation steps are respectively replaced by successive mapping and fusion processes. The main steps of this methodology consist in (figure ~\ref{fig:DissociationGB}) ($1$) analyzing the decision problem through a hierarchical structure, ($2$) defining the evaluation classes for decision through a common frame of discernment, ($3$) evaluating the qualitative or quantitative criteria, ($4$) mapping the evaluations of criteria into the common frame of discernment for decision, ($5$) fusing the mapped evaluations of criteria to get a basic belief assignment related to the evaluations classes of decision (frame of discernment for decision). These steps are independent one from each other. Therefore, imprecise and uncertain evaluations of quantitative or qualitative criteria can be done by the sources (experts) and re-used with different mapping models.

        \subsection{Decision problem elicitation}

\begin{figure}[htb]
        \begin{center}
        \includegraphics[width=88mm]{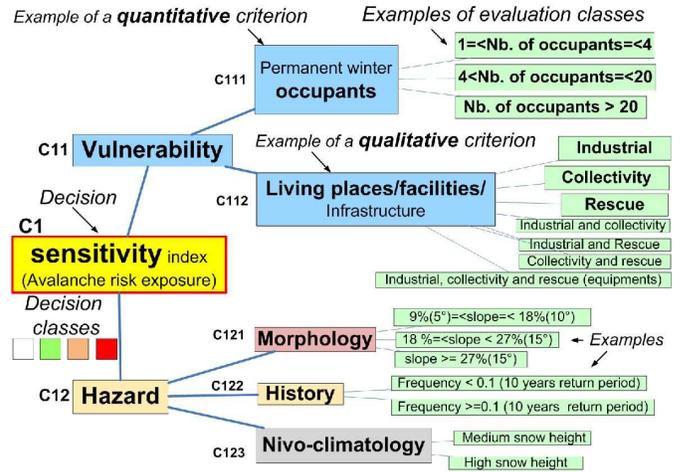}
        \end{center}
        \caption{A sample decision problem.}
        \label{fig:ArbreSimpleGB}
    \end{figure}

     The methodology is described through a sample problem dealing with the sensitivity of an avalanche prone area (figure~\ref{fig:ArbreSimpleGB}) derived from a real existing decision framework used to identify the sensitive avalanche paths ~\cite{Rapin2006}. For each avalanche path, decision consists in choosing between sensitivity levels described as not, low, medium or high sensitive paths. This sensitivity depends on both vulnerability and hazard levels. Vulnerability is assessed through the number of winter permanent occupants (quantitative criterion) and existing infrastructures (qualitative criterion).

\begin{figure}[htb]
        \begin{center}
        \includegraphics[width=88mm]{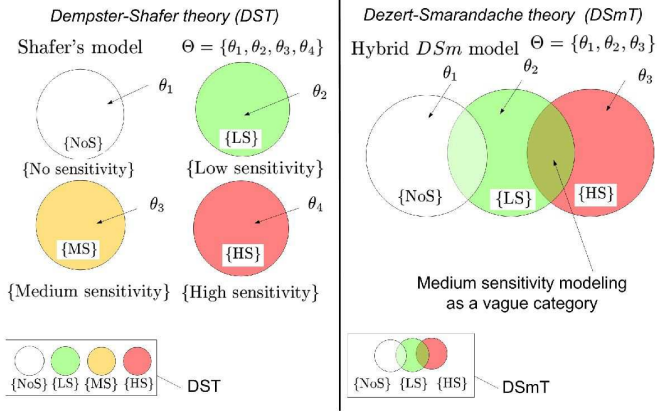}
        \end{center}
        \caption{Two frames of discernment according to $DST$ or $DSmT$ frameworks.}
        \label{fig:FrameDiscernmentDSTDsmT}
    \end{figure}

     Sensitivity levels are the elements of the common frame of discernment for decision. The fusion process will provide basic belief assignments on each or combination of the elements of this frame of discernment. In the classical $DST$ framework based on exhaustive and exclusive hypotheses,
     the frame $\Theta$ is composed of $4$ exclusive elements defined by $HD_{1}=\text{'No sensitivity'}$, $HD_{2}=\text{'Low sensitivity'}$, $HD_{3}=\text{'Medium sensitivity'}$ and $HD_{4}=\text{'High sensitivity'}$. In the $DSmT$ framework (allowing non-empty intersections), the frame $\Theta$ is composed of $3$ elements defined by $HD_{1}=\text{'No sensitivity'}$, $HD_{2}=\text{'Low sensitivity'}$ and $HD_{3}=\text{'High sensitivity'}$ (figure~\ref{fig:FrameDiscernmentDSTDsmT}).\\

        \subsection{Evaluation of quantitative criteria}

\begin{figure}[htb]
        \begin{center}
        \includegraphics[width=88mm]{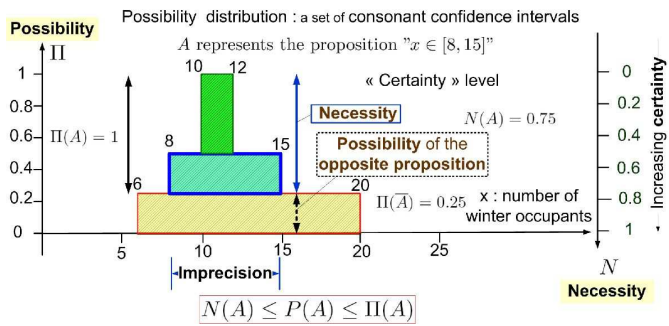}
        \end{center}
        \caption{Using possibility distribution for imprecise criteria evaluation.}
        \label{fig:PrincipePossibiliteLTGB}
    \end{figure}

    The quantitative criterion are evaluated through possibility distributions which allow to represent both imprecision and uncertainty. The source (an expert) provides evaluations as intervals. Let us take as an example the criterion $C_{111}$ corresponding to the number of permanent winter occupants: $A$ represents the proposition "$x\in [8,15]$". $N(A)=0.75$ represents the certainty level (confidence) in the proposition "$x\in [8,15]$". $N(A)$ can be viewed as a lowest probability for $A$ and $\Pi(A)$ as un upper probability for $A$  (figure~\ref{fig:PrincipePossibiliteLTGB}).

\begin{figure}[h]
        \begin{center}
        \includegraphics[width=88mm]{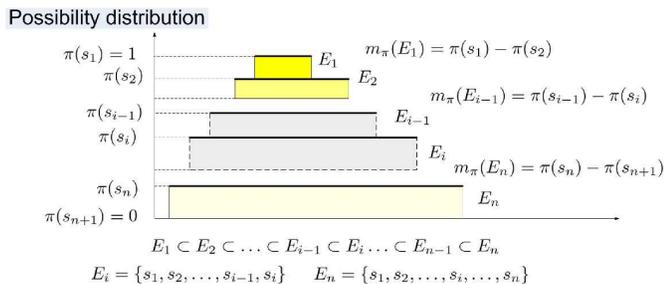}
        \end{center}
        \caption{Possibility distribution for evaluation are transformed into masses.}
        \label{fig:PossibilityToMassesGB}
    \end{figure}

      Any interval of the possibility distribution can be transformed into basic belief assignment (figure~\ref{fig:PossibilityToMassesGB}) according to relations between possibility and belief function theories ~\cite{Baudrit2005,Baudrit2005a,Dubois2006}.

        \subsection{Mapping models: a link from evaluation to decision}

    \begin{figure}[h]
        \begin{center}
        \includegraphics[width=88mm]{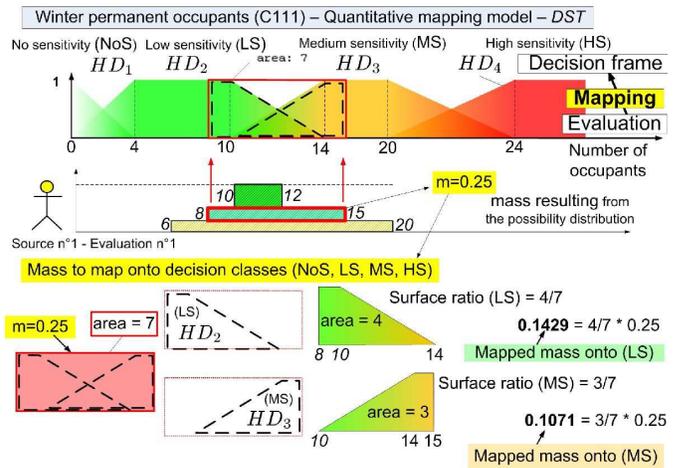}
        \end{center}
        \caption{Mapping is based on surface ratios.}
        \label{fig:SurfaceRatioSimpleGB}
    \end{figure}

\begin{figure}[h]
        \begin{center}
        \includegraphics[width=88mm]{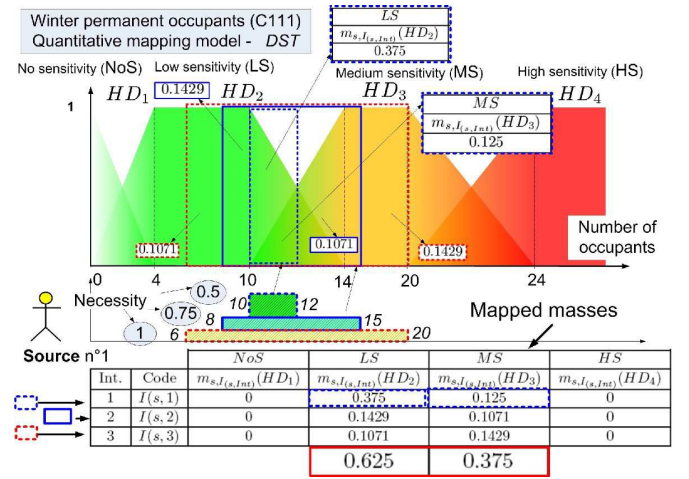}
        \end{center}
        \caption{Results of mapping of the evaluation of source n°1 for criterion $C_{111}$.}
        \label{fig:RecapMappingGB}
    \end{figure}

     A mapping model is a set of fuzzy intervals $L-R$ linking a criterion evaluation and the decision classes: it plays more or less the same role than the utility function in a total aggregation based multi-criteria decision method. For each evaluation of a criterion by one source, each interval of the possibility distribution ($I_{(s,int_{j})}$) is mapped to the common frame of discernment of decision  according to surface ratios (figures ~\ref{fig:SurfaceRatioSimpleGB}, ~\ref{fig:RecapMappingGB}). At the end of the mapping process, all the criteria evaluations provided by each source are transformed in basic belief assignments (bba's) according the common frame of discernment of decision: these bba's are then fused in a two-step process.

        \subsection{Two steps of fusion}

\begin{figure}[h]
        \begin{center}
        \includegraphics[width=88mm]{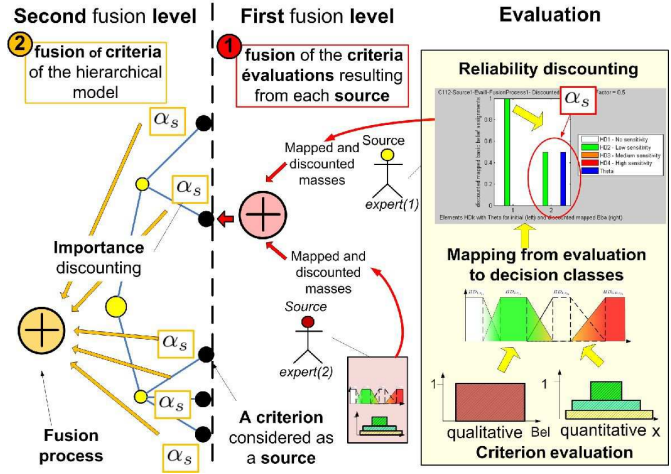}
        \end{center}
        \caption{The fusion levels of the $ER-MCDA$ process.}
        \label{fig:EtapeProcessSimpleGB}
    \end{figure}

    After the mapping step, the $ER-MCDA$ process is based on two successive fusion levels (figure~\ref{fig:EtapeProcessSimpleGB}). The first step consists in the fusion of bba's corresponding, for each criterion, to the different evaluations provided by different sources.

\begin{figure}[h]
        \begin{center}
        \includegraphics[width=88mm]{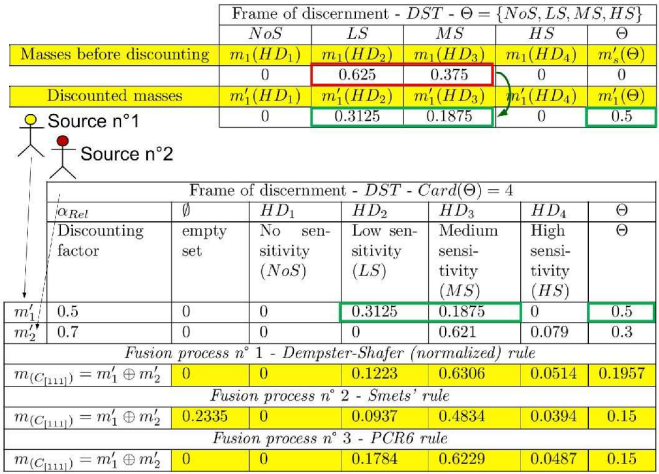}
        \end{center}
        \caption{Examples of fusion results.}
        \label{fig:Tab9.7et9.3}
   \end{figure}

    Reliability of each source is assessed using the classical discounting factor ($\alpha_{s}$) proposed in the $DST$ or $DSmT$ frameworks. Then, fusion of this discounted bba's is done using different fusion rules. The $PCR6$ rule ~\cite{Dezert2006a,Dezert2009,Martin2006} is recommended to prevent aberrant decisions in case of highly conflicting evaluations of a criterion (figure~\ref{fig:Tab9.7et9.3}).

\begin{figure}[h]
        \begin{center}
        \includegraphics[width=88mm]{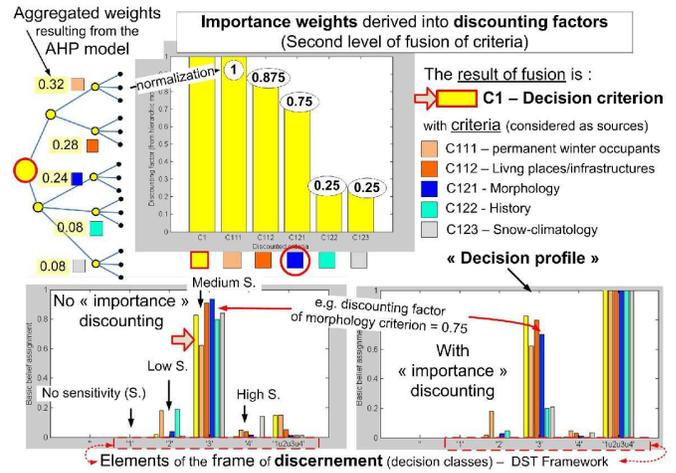}
        \end{center}
        \caption{Importance discounting factors resulting from hierarchical multi-criteria model.}
        \label{fig:compAffaiblissementGB}
   \end{figure}

   The second step consists in the fusion of the bba's corresponding to each criterion and resulting from the first step of fusion (figure~\ref{fig:EtapeProcessSimpleGB}). In this second step, each criterion is considered as a source which is discounted according its importance in the decision process as proposed by ~\cite{Beynon2000} (figure~\ref{fig:compAffaiblissementGB}).

\begin{figure}[h]
        \begin{center}
        \includegraphics[width=88mm]{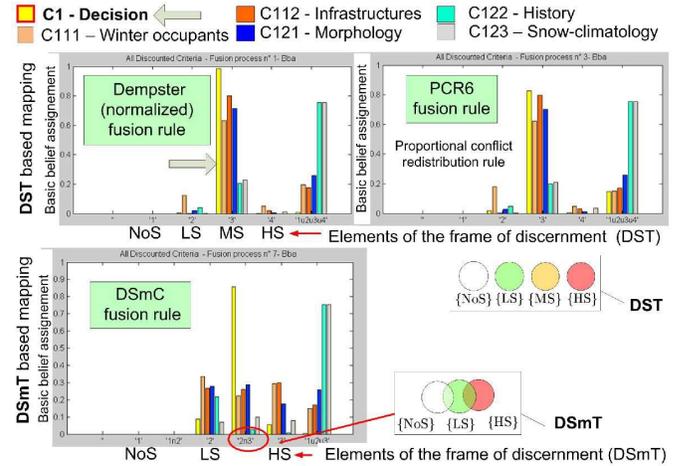}
        \end{center}
        \caption{Comparison of fusion rules.}
        \label{fig:CompDempsterPCR6DSmC}
   \end{figure}

    Results of fusion have to be interpreted to decide which is the sensitivity level that will be chosen ($NoS$, $LS$, $MS$ or $HS$) according either to the maximum of basic belief assignments, credibility (pessimistic decision), plausibility (optimistic decision) or pignistic probability (compromise). The $ER-MCDA$ methodology produces a comparative decision profile in which decision classes (elements of the frame of discernment) can be compared one to each other using $DST$ or $DSmT$ to fit in the best possible way to their nature (figure~\ref{fig:CompDempsterPCR6DSmC}).

    \section{Conclusion-Discussion}
    \label{sec:Conclusion}

    The $ER-MCDA$ methodology allows to make a decision based on multiple and more or less important criteria on which more or less reliable sources provide imperfect and uncertain evaluations. A simplified decision sorting problem based on a snow-avalanche risk management problem shows how the use of multi-criteria decision analysis principles and information fusion can be used to characterize and take information quality or imperfection into account for decision purposes. In regard with its aggregation principles and possible "rank reversals", $AHP$ is as much critized than it is widely used ~\cite{Linkov2006}. Anyway, it remains an easy understandable method that can be simply connected to fusion process frameworks. The Analytic Hierarchy Process ($AHP$) elicits the criteria used for decision and is used as a conceptual framework. The $ER-MCDA$ methodology contributes to improve traceability and quality description of the expertise process through clearly dissociated steps corresponding respectively to evaluation, mapping and fusion based decision making. $DSmT$ proposes more valuable modeling principles for vague, imprecise and uncertain information and  conflict management. Advanced fusion rules such as partial conflicting rules ($PCR$) cope with conflict in a more efficient way than the classical Dempster's rule used in the $DST$ framework.

    Sensitivity analysis must still be applied to the $ER-MCDA$ methodology in order to explore the effects of fusion orders, mapping models \ldots changes. Using the classical discounting factor to consider both reliability and uncertainty at the first fusion step and importance at the second step of fusion is not satisfactory. A new discounting process must be proposed in this case ~\cite{Tacnet2009a, Tacnet2009b, Dezert2010}.

    From an operational point of view, an important application field consists in extending this methodology to spatial applications and specially to hazard and risk zoning maps.

    \section*{Acknowledgements}

This work would not have been possible without the help of Arnaud Martin who provided us the fusion routines that were used to implement our calculation framework ~\cite{Martin2009}. 


\bibliographystyle{IEEEtran}

\begin{thebibliography}{1}
\providecommand{\url}[1]{#1}
\csname url@rmstyle\endcsname
\providecommand{\newblock}{\relax}
\providecommand{\bibinfo}[2]{#2}
\providecommand\BIBentrySTDinterwordspacing{\spaceskip=0pt\relax}
\providecommand\BIBentryALTinterwordstretchfactor{4}
\providecommand\BIBentryALTinterwordspacing{\spaceskip=\fontdimen2\font plus
\BIBentryALTinterwordstretchfactor\fontdimen3\font minus
  \fontdimen4\font\relax}
\providecommand\BIBforeignlanguage[2]{{%
\expandafter\ifx\csname l@#1\endcsname\relax
\else
\language=\csname l@#1\endcsname
\fi
#2}}


%
%



\bibitem{Baudrit2005}
C. Baudrit," Représentation et propagation de connaissances imprécises et
  incertaines : application à l'évaluation des risques liés aux sites et aux
  sols pollués ", PhD thesis, Toulouse III University -U.F.R. Mathématiques
  Informatique Gestion, Toulouse, 2005.

\bibitem{Baudrit2005a}
C. Baudrit, D. Guyonnet, and D. Dubois, "Postprocessing the hybrid method for addressing uncertainty in risk
  assessments", \emph{Journal of Environmental Engineering},vol.~131, no.~12, pp. 1750--1754, 2005.


\bibitem{Beynon2000}
M. Beynon, B. Curry, and P. Morgan, "The Dempster-Shafer theory of evidence:an alternative approach to
  multicriteria decision modelling", \emph{Omega}, vol.~28, no.~1, pp. 37--50, 2000.

\bibitem{Beynon2002}
M. Beynon, "DS/AHP method:a mathematical analysis, including an understanding of
  uncertainty", \emph{European Journal of Operational Research},vol.~140, no.~1, pp.148--164, 2002.


\bibitem{Beynon2005a}
M. Beynon, "A method of aggregation in ds/ahp for group decision-making with the
  non-equivalent importance of individuals in the group",\emph{Computers and Operations Research}, vol.~32, pp.1881--1896, 2005.

\bibitem{Dezert2006a}
J. Dezert and F. Smarandache, "Proportional
  Conflict Redistribution Rules for Information Fusion" in \emph{Advances and applications of DSmT for Information Fusion-
  Collected works - Volume 2 - Dezert J., Smarandache F. (Eds.) - American Research Press, Rehoboth, USA.}, pp. 3--68, 2006.

\bibitem{Dezert2009}
J. Dezert and F. Smarandache, "An introduction to DSmT" in \emph{Advances and applications of DSmT for Information Fusion -
  Collected works - Volume 3 - Dezert J., Smarandache F. (Eds.) - American Research Press, Rehoboth, USA.}, pp. 3--73, 2009.

\bibitem{Dezert2010}
J. Dezert, F. Smarandache, J.M. Tacnet and M. Batton-Hubert, "Multi-criteria decision making based on DSmT/AHP", \emph{submitted to International Workshop on Belief Functions}, Brest, France, April 2010.



\bibitem{Dubois1988}
D. Dubois and H. Prade, "Possibility Theory:an approach to Computerized Processing of
  Uncertainty", Plenum Press, New York (U.S.A), 1988.

\bibitem{Dubois2006}
D. Dubois and H. Prade, "Représentations formelles de l'incertain et de
  l'imprécis", in
\emph{Concepts et méthodes pour l'aide à la décision - Volume 1 : outils de
  modélisation, Bouyssou, D., Dubois, D., Pirlot, M., and Prade, H. (Eds.), Hermès-Lavoisier, Paris, 2006}.

\bibitem{Dyer2005}
J. Dyer, "MAUT - Multiattribute Utility theory" in \emph{Multiple Criteria Decision Analysis:State of the Art Surveys - Figueira J., Greco S., Ehrgott, M. (Eds.) - Springer Verlag, Boston, Dordrecht, London}, vol.~78 of International Series in Operations Research and Management Science, pp.263--295, 2005.

\bibitem{Forman2002}
E. Forman and M.~A. Selly , "Decision by Objectives", World Scientific Publishing, Singapore, 2002.

\bibitem{Keeney1976}
R. Keeney and H. Raiffa, "Decisions with multiple objectives:preferences and values
  trade-offs", J.Wiley and Sons, New York, 1976.

\bibitem{Linkov2006}
I. Linkov, F. Satterstrom, G. Kiker, C. Batchelor, T. Bridges and E. Ferguson
  Ferguson, E., "From comparative risk assessment to multi-criteria decision analysis
  and adaptive management:recent developments and applications", \emph{Environment International Environmental Risk Management - the
  State of the Art}, vol.~32, no.~8, pp.1072--1093, 2006.

\bibitem{Martin2006}
A. Martin and C. Osswald, "A new generalization of the proportional
  conflict distribution rule stable in terms of decision" in \emph{Advances and applications of DSmT for Information Fusion-
  Collected works - Volume 2 - Dezert J., Smarandache F. (Eds.) - American Research Press, Rehoboth, USA.}, pp. 69--88, 2006.


\bibitem{Martin2009}
A. Martin, "Implementing general belief function
  framework with a practical codification for low complexity" in \emph{Advances and applications of DSmT for Information Fusion -
  Collected works - Volume 3 - Dezert J., Smarandache F. (Eds.) - American Research Press, Rehoboth, USA.}, pp. 217--273, 2009.

\bibitem{Omrani2007}
H. Omrani, L. Ion-Boussier and P. Trigano, "A new approach for impacts assessment of urban mobility",
\emph{WSEAS Transactions on Information Science and Applications},vol.~4, no.~3, pp. 439--444, 2007.

\bibitem{Rapin2006}
F. Rapin, L. Belanger, A. Hurand and J.M. Bernard, "Sensitive avalanche paths:using a new method for inventory and
  classification of risk" in \emph{International Snow Science Workshop Proceedings (ISSW) 2006},
  september 30 to october 6, Telluride, Colorado, United States, 2006.


\bibitem{Saaty1980}
T. Saaty, "The analytic hierarchy process", McGraw Hill, New York, 1980.

\bibitem{Saaty2005}
T. Saaty, "The analytic hierarchy and analytic network processes for the measurement of intangible criteria and for decision making" in \emph{Multiple Criteria Decision Analysis:state of the Art Surveys - Figueira J., Greco S., Ehrgott, M. (Eds.) - Springer Verlag, Boston, Dordrecht, London}, vol.~78 of International Series in Operations Research and Management Science, pp.345--407, 2005.

\bibitem{Shafer1976}
G. Shafer, "A mathematical theory of evidence", Princeton University Press, 1976.

\bibitem{DSmTBook1-3}
F. Smarandache and J. Dezert, "Advances and applications of DSmT for information fusion (Collected works)", Vol. 1-3, American Research Press, 2004--2009.
{\small{http://www.gallup.unm.edu/{\verb+~+}smarandache/DSmT.htm}}

\bibitem{Roy1989}
B. Roy, "Main sources of inaccurate determination, uncertainty and imprecision
  in decision models", \emph{Mathematical and Computer Modelling}, vol.~12, no.~10-11, pp. 1245--1254, 1989.

\bibitem{Stewart2005a}
T.~J. Stewart , "Dealing with uncertainties in MCDA" in \emph{Multiple Criteria Decision Analysis:state of the Art Surveys - Figueira J., Greco S., Ehrgott, M. (Eds.) - Springer Verlag, Boston, Dordrecht, London}, vol.~78 of International Series in Operations Research and Management Science, pp.445--470, 2005.

\bibitem{Tacnet2009a}
J.M. Tacnet, M. Batton-Hubert and J. Dezert, "Information fusion for natural hazards
  in mountains" in \emph{Advances and applications of DSmT for Information Fusion-
  Collected works - Volume 3 - Dezert J., Smarandache F. (Eds.) - American Research Press, Rehoboth, USA.}, pp. 565--659, 2009.

\bibitem{Tacnet2009b}
J.M. Tacnet, "Prise en compte de l'incertitude dans l'expertise des risques
  naturels en montagne par analyse multicritères et fusion d'information", Phd in \emph{Sciences et génie de l'environnement}, Ecole Nationale Supérieure des
  Mines, Saint-Etienne, France, 2009.

\bibitem{Vaidya2006}
O.~S. Vaidya and S. Kumar, "Analytic hierarchy process:an overview of applications", \emph{European Journal of Operational Research}, vol.~169, no.~1, pp. 1--29, 2006.

\bibitem{Wang2006b}
Y.~M. Wang, J.~B. Yang, and D.~L. Xu, "Environmental impact assessment using the evidential reasoning
  approach", \emph{European Journal of Operational Research}, vol.~174, no.~3, pp. 1885--1913, 2006.

\bibitem{Zadeh1965}
L. Zadeh, "Fuzzy sets", \emph{Information and Control}, vol.~8, pp. 338--353, 1965.

\bibitem{Zadeh1978}
L. Zadeh, "Fuzzy sets as a basis for a theory of possibility", \emph{Fuzzy Sets and Systems}, vol.~1 , pp. 3--28, 1978.







\end{thebibliography}

\end{document}